\PassOptionsToPackage{unicode}{hyperref}
\PassOptionsToPackage{hyphens}{url}
\PassOptionsToPackage{dvipsnames,svgnames,x11names}{xcolor}
\documentclass[
]{article}
\usepackage{xcolor}
\usepackage{amsmath,amssymb}
\usepackage{iftex}
\usepackage[T1]{fontenc}
\usepackage[utf8]{inputenc}
\usepackage{textcomp}
\IfFileExists{upquote.sty}{\usepackage{upquote}}{}
\IfFileExists{microtype.sty}{
  \usepackage[]{microtype}
  \UseMicrotypeSet[protrusion]{basicmath} 
}{}
\makeatletter
\@ifundefined{KOMAClassName}{
  \IfFileExists{parskip.sty}{%
    \usepackage{parskip}
  }{
    \setlength{\parindent}{0pt}
    \setlength{\parskip}{6pt plus 2pt minus 1pt}}
}{
  \KOMAoptions{parskip=half}}
\makeatother
\makeatletter
\ifx\paragraph\undefined\else
  \let\oldparagraph\paragraph
  \renewcommand{\paragraph}{
    \@ifstar
      \xxxParagraphStar
      \xxxParagraphNoStar
  }
  \newcommand{\xxxParagraphStar}[1]{\oldparagraph*{#1}\mbox{}}
  \newcommand{\xxxParagraphNoStar}[1]{\oldparagraph{#1}\mbox{}}
\fi
\ifx\subparagraph\undefined\else
  \let\oldsubparagraph\subparagraph
  \renewcommand{\subparagraph}{
    \@ifstar
      \xxxSubParagraphStar
      \xxxSubParagraphNoStar
  }
  \newcommand{\xxxSubParagraphStar}[1]{\oldsubparagraph*{#1}\mbox{}}
  \newcommand{\xxxSubParagraphNoStar}[1]{\oldsubparagraph{#1}\mbox{}}
\fi
\makeatother

\usepackage{longtable,booktabs,array}
\usepackage{calc} 
\usepackage{etoolbox}
\makeatletter
\patchcmd\longtable{\par}{\if@noskipsec\mbox{}\fi\par}{}{}
\makeatother
\IfFileExists{footnotehyper.sty}{\usepackage{footnotehyper}}{\usepackage{footnote}}
\makesavenoteenv{longtable}
\usepackage{graphicx}
\makeatletter
\newsavebox\pandoc@box
\newcommand*\pandocbounded[1]{
  \sbox\pandoc@box{#1}%
  \Gscale@div\@tempa{\textheight}{\dimexpr\ht\pandoc@box+\dp\pandoc@box\relax}%
  \Gscale@div\@tempb{\linewidth}{\wd\pandoc@box}%
  \ifdim\@tempb\p@<\@tempa\p@\let\@tempa\@tempb\fi
  \ifdim\@tempa\p@<\p@\scalebox{\@tempa}{\usebox\pandoc@box}%
  \else\usebox{\pandoc@box}%
  \fi%
}
\def\fps@figure{htbp}
\makeatother

\NewDocumentCommand\citeproctext{}{}
\NewDocumentCommand\citeproc{mm}{%
  \begingroup\def\citeproctext{#2}\cite{#1}\endgroup}
\makeatletter
 \let\@cite@ofmt\@firstofone
 \def\@biblabel#1{}
 \def\@cite#1#2{{#1\if@tempswa , #2\fi}}
\makeatother
\newlength{\cslhangindent}
\newlength{\csllabelwidth}
\newenvironment{CSLReferences}[2] 
 {\begin{list}{}{%
  \setlength{\itemindent}{0pt}
  \setlength{\leftmargin}{0pt}
  \setlength{\parsep}{0pt}
  \ifodd #1
   \setlength{\leftmargin}{\cslhangindent}
   \setlength{\itemindent}{-1\cslhangindent}
  \fi
  \setlength{\itemsep}{#2\baselineskip}}}
 {\end{list}}
\usepackage{calc}

\providecommand{\tightlist}{%
  \setlength{\itemsep}{0pt}\setlength{\parskip}{0pt}}

\usepackage{booktabs}
\usepackage{caption}
\usepackage{longtable}
\usepackage{colortbl}
\usepackage{array}
\usepackage{anyfontsize}
\usepackage{multirow}
\usepackage{tcolorbox}
\usepackage{arxiv}
\usepackage{orcidlink}
\usepackage{amsmath}
\usepackage[T1]{fontenc}
\makeatletter
\@ifpackageloaded{caption}{}{\usepackage{caption}}
\AtBeginDocument{%
\ifdefined\contentsname
  \renewcommand*\contentsname{Table of contents}
\else
  \newcommand\contentsname{Table of contents}
\fi
\ifdefined\listfigurename
  \renewcommand*\listfigurename{List of Figures}
\else
  \newcommand\listfigurename{List of Figures}
\fi
\ifdefined\listtablename
  \renewcommand*\listtablename{List of Tables}
\else
  \newcommand\listtablename{List of Tables}
\fi
\ifdefined\figurename
  \renewcommand*\figurename{Figure}
\else
  \newcommand\figurename{Figure}
\fi
\ifdefined\tablename
  \renewcommand*\tablename{Table}
\else
  \newcommand\tablename{Table}
\fi
}
\@ifpackageloaded{float}{}{\usepackage{float}}
\floatstyle{ruled}
\@ifundefined{c@chapter}{\newfloat{codelisting}{h}{lop}}{\newfloat{codelisting}{h}{lop}[chapter]}
\floatname{codelisting}{Listing}

\makeatother
\makeatletter
\@ifpackageloaded{caption}{}{\usepackage{caption}}
\@ifpackageloaded{subcaption}{}{\usepackage{subcaption}}
\makeatother
\usepackage{bookmark}
\IfFileExists{xurl.sty}{\usepackage{xurl}}{} 
\hypersetup{
  pdftitle={Evaluating Large Language Models for IUCN Red List Species Information},
  pdfauthor={Shinya Uryu},
  pdfkeywords={Large language models, IUCN Red List, Biodiversity
informatics, Conservation assessment, Artificial intelligence},
  colorlinks=true,
  linkcolor={blue},
  filecolor={Maroon},
  citecolor={Blue},
  urlcolor={Blue},
  pdfcreator={LaTeX via pandoc}}

\newcommand{\runninghead}{A Preprint }
\renewcommand{\runninghead}{A Preprint }
\title{Evaluating Large Language Models for IUCN Red List Species
Information}
\author{\textbf{Shinya Uryu}~\orcidlink{0000-0002-0493-6186}\\Center for
Design-Oriented AI Education and Research\\Tokushima
University\\Tokushima,\ 770-8502\\\href{mailto:uryu.shinya@tokushima-u.ac.jp}{uryu.shinya@tokushima-u.ac.jp}}
\date{}
\begin{document}
\maketitle
\begin{abstract}
Large Language Models (LLMs) are rapidly being adopted in conservation
to address the biodiversity crisis, yet their reliability for species
evaluation is uncertain. This study systematically validates five
leading models on 21,955 species across four core IUCN Red List
assessment components: taxonomy, conservation status, distribution, and
threats. A critical paradox was revealed: models excelled at taxonomic
classification (94.9\%) but consistently failed at conservation
reasoning (27.2\% for status assessment). This knowledge-reasoning gap,
evident across all models, suggests inherent architectural constraints,
not just data limitations. Furthermore, models exhibited systematic
biases favoring charismatic vertebrates, potentially amplifying existing
conservation inequities. These findings delineate clear boundaries for
responsible LLM deployment: they are powerful tools for information
retrieval but require human oversight for judgment-based decisions. A
hybrid approach is recommended, where LLMs augment expert capacity while
human experts retain sole authority over risk assessment and policy.
\end{abstract}
{\bfseries \emph Keywords}
\def\sep{\textbullet\ }
Large language models \sep IUCN Red List \sep Biodiversity
informatics \sep Conservation assessment \sep 
Artificial intelligence

\section{Introduction}\label{sec-intro}

Global biodiversity faces a severe crisis, with extinction rates
estimated to be 100--1,000 times higher than background levels
(\citeproc{ref-devos2015}{De Vos et al. 2015};
\citeproc{ref-cowie2022}{Cowie, Bouchet, and Fontaine 2022}). As of
March 2025, the IUCN Red List has assessed over 169,000 species---yet
this represents fewer than 2\% of the estimated 8 million species
worldwide (\citeproc{ref-borgelt2022}{Borgelt et al. 2022}). Moreover,
56\% of ``Data Deficient'' species may actually be threatened,
highlighting systematic gaps in the knowledge base for conservation
policy (\citeproc{ref-borgelt2022}{Borgelt et al. 2022};
\citeproc{ref-cazalis2022}{Cazalis et al. 2022};
\citeproc{ref-finn2023}{Finn, Grattarola, and Pincheira-Donoso 2023}).
Assessment coverage remains heavily biased toward vertebrates and Global
North species, while invertebrates and tropical taxa are considerably
underrepresented (\citeproc{ref-cazalis2022}{Cazalis et al. 2022}).

Red List evaluations are designed to be rigorous, relying on expert
judgment and detailed data collection. However, the process is time- and
resource-intensive, often requiring 2--5 years per major taxonomic group
(IUCN 2025). Combined with geographic and taxonomic biases in available
expertise, this leads to delays that hinder timely conservation action.
A persistent gap therefore remains between academic biodiversity
research and its incorporation into conservation assessments
(\citeproc{ref-cazalis2022}{Cazalis et al. 2022}).

Computational approaches have been proposed to address this scalability
challenge. Machine learning models can predict extinction risk
categories for unevaluated taxa, offering pathways to expand coverage
(\citeproc{ref-zizka2021}{Zizka et al. 2021};
\citeproc{ref-caetano2022}{Caetano et al. 2022};
\citeproc{ref-borgelt2022}{Borgelt et al. 2022};
\citeproc{ref-lucas2024}{Lucas et al. 2024}). However, such models
typically rely on structured datasets and cannot easily incorporate the
heterogeneous, text-based evidence that underpins Red List assessments.
Large language models (LLMs) represent a qualitatively different
opportunity: they can synthesize unstructured information across domains
and communicate results in natural language, potentially democratizing
access to conservation knowledge (\citeproc{ref-cooper2024}{Cooper et
al. 2024}; \citeproc{ref-dagdelen2024}{Dagdelen et al. 2024};
\citeproc{ref-keck2025}{Keck, Broadbent, and Altermatt 2025}).

At the same time, their propensity for ``hallucinations''---fluent but
fabricated outputs---introduces new risks
(\citeproc{ref-farquhar2024}{Farquhar et al. 2024};
\citeproc{ref-mammides2024}{Mammides and Papadopoulos 2024};
\citeproc{ref-huang2025}{Huang et al. 2025}). These dual
potentials---accelerating access versus amplifying error---necessitate
rigorous evaluation. While a growing body of research has begun to
examine LLM performance in ecological contexts
(\citeproc{ref-farrell2024}{Farrell et al. 2024};
\citeproc{ref-gougherty2024}{Gougherty and Clipp 2024}), systematic
benchmarking remains nascent. Dorm et al.
(\citeproc{ref-dorm2025}{2025}) recently pioneered this field by
evaluating Gemini 1.5 Pro and GPT-4o across ecological tasks, finding
that while LLMs outperformed naive baselines, they exhibited
``significant limitations, particularly in generating spatially accurate
range maps and classifying threats''---precisely the capabilities most
critical for conservation assessment.

To address this gap, this study presents the first comprehensive,
multi-model evaluation of LLM reliability, assessing five
state-of-the-art models across 21,955 species reassessed in
2022--2023---nearly ten times the scale of previous studies. My analysis
addresses three overarching questions: (1) How accurately do LLMs
reproduce IUCN Red List information across different task types? (2) How
do performance patterns differ across taxonomic groups? (3) What
mechanisms explain the observed perforance differences between
information retrieval and ecological reasoning tasks?

Given the well-documented biases in conservation research and data
availability, I also examine whether LLMs amplify existing taxonomic
inequities. Specifically, I investigate: (i) whether LLMs show
systematic performance differences between major taxonomic groups
(vertebrates vs.~invertebrates vs.~plants vs.~fungi); (ii) within
vertebrates, whether traditional conservation priority groups (mammals,
birds) demonstrate superior performance compared to less well-studied
taxa (amphibians, reptiles, fishes); and (iii) whether taxonomic
performance differences vary by task type, suggesting task-specific
knowledge biases. These questions are critical because any deployment of
LLMs in conservation must account for---and ideally correct---systematic
biases that could further marginalize already understudied taxa.

By answering these questions, I establish the most systematic benchmark
to date of LLM reliability in conservation contexts. My findings clarify
both opportunities and risks: LLMs may be valuable for education, public
engagement, and exploratory data retrieval, but remain unsuitable for
conservation assessments, threat prioritization, or policy use without
expert validation. This study not only benchmarks current capabilities
but also delineates a roadmap for responsibly integrating LLMs with
expert-curated resources such as the IUCN Red List.

\section{Methods}\label{sec-methods}

\subsection{Target Species and Reference Dataset}\label{subsec-2-1}

This study analyzed 21,962 assessment records for 21,955 unique species
evaluated by the IUCN Red List between 2022 and 2023, a dataset that
included multiple assessments for seven species.

The dataset comprised vertebrates (10,575 species, 48.17\%), plants
(9,800, 44.64\%), invertebrates (1,353, 6.16\%), fungi (224, 1.02\%),
and chromista (3, 0.01\%). The Chromista group, comprising only 3
species, was retained in descriptive analyses but excluded from all
formal statistical comparisons due to its insufficient sample size.
Among vertebrates, fishes were the most numerous (5,055, 23.02\%),
followed by amphibians (3,187, 14.52\%), birds (1,223, 5.57\%), reptiles
(917, 4.18\%), and mammals (193, 0.88\%). This distribution reflects
IUCN's historical emphasis on vertebrates and highlights the relative
underrepresentation of invertebrates and fungi.

All major IUCN Red List categories were represented, including
Critically Endangered (CR), Endangered (EN), Vulnerable (VU), Near
Threatened (NT), Least Concern (LC), and Data Deficient (DD)
(\citeproc{ref-iucn_guidelines_2024}{IUCN Standards and Petitions
Committee 2024}). Reference data were obtained from the IUCN Red List
API (version 4, accessed August 2025 IUCN
(\citeproc{ref-iucn_api_2025}{2025b})), which provided standardized
information on taxonomy, conservation status, distribution, and threats.
Reproducibility was ensured by version pinning, and categories
unavailable via the API (Not Evaluated, NE; Not Recognized, NR) were
excluded.

\subsection{Large Language Models
(LLMs)}\label{large-language-models-llms}

In August 2025, five state-of-the-art LLMs were evaluated: three
API-based systems---GPT-4.1 by OpenAI, Grok 3 by xAI, and Claude Sonnet
4 by Anthropic---and two locally hosted open-weight models executed via
Ollama, Gemma 3-27B by Google DeepMind and Llama 3.3-70B by Meta.
Reasoning-specialized variants (e.g., o3, o4-mini) were excluded, since
the study focused on general knowledge accuracy.

These models differ in several respects: GPT-4.1 and Claude Sonnet 4
rely on fixed knowledge cutoffs, Grok 3 supplements responses with
real-time web search, and the open-weight models Gemma 3-27B and Llama
3.3-70B emphasize efficiency and broad accessibility. They also span
different deployment environments (cloud API vs.~local execution) and
parameter scales (27B--70B for open-weight models; undisclosed for
proprietary systems). To preserve comparability despite this
heterogeneity, all models were assessed under uniform conditions using
minimal prompting with standardized formatting, rather than through
system-specific prompt optimization.

\begin{table}

\caption{\label{tbl-1}Summary of large language models included in the
comparative assessment.}

\centering{

\fontsize{9.0pt}{10.8pt}\selectfont
\begin{tabular*}{\linewidth}{@{\extracolsep{\fill}}llrlr}
\toprule
Provider & Model & Knowledge reference date & Type & Context window (tokens) \\ 
\midrule\addlinespace[2.5pt]
OpenAI & GPT-4.1 & 2024-06 & Proprietary & 1,047,576 \\ 
Anthropic & Claude Sonnet 4 & 2025-01 & Proprietary & 200,000 \\ 
xAI & Grok 3 & NA & Proprietary & 131,072 \\ 
Ollama & gemma3:27b & 2024-08 & Open-weight & 131,072 \\ 
Ollama & llama3.3:70b & 2023-12 & Open-weight & 131,072 \\ 
\bottomrule
\end{tabular*}
\begin{minipage}{\linewidth}
Grok 3 has web search capability, but this feature was disabled during evaluation to ensure comparability. Knowledge reference date is not applicable as the system does not maintain a fixed training cutoff.\\
\end{minipage}

}

\end{table}%

\subsection{Evaluation Tasks and
Metrics}\label{evaluation-tasks-and-metrics}

Task 1 (taxonomic classification) employed zero-shot prompting without
input-output examples. For Tasks 2--4, we embedded 2--3 representative
input-output pairs within the system instructions to clarify expected
response formats (see Appendix A). All prompts were delivered as system
instructions, which ensured consistent task interpretation and
reproducibility across API-based and locally-hosted models. The first
task required taxonomic classification from Kingdom to Family level
using multiple-choice questions, with two variants: Task 1a employed
phylogenetically similar distractors generated from the GBIF Backbone
Taxonomy (\citeproc{ref-GBIF2023}{GBIF Secretariat 2023}) using
taxonomic distinctness metrics (\citeproc{ref-clarke1998}{K. R. Clarke
and Warwick 1998}; \citeproc{ref-clarke2001}{K. Clarke and Warwick
2001}), while Task 1b used randomly selected distractors from the same
taxonomic level. The second task required models to identify IUCN Red
List conservation status categories. The third task evaluated geographic
distribution knowledge by requiring models to determine species ranges
at the country level from a standardized list of 195 ISO-3166-1 country
names (\citeproc{ref-iucn_country_names}{IUCN 2025a}). The fourth task
assessed threat identification, requiring models to select applicable
threat categories from the 12 Level-1 IUCN Threat Categories (Version
3.3 IUCN (\citeproc{ref-iucn_threats_classification_scheme}{2025c})).
Species lacking threat data were excluded only from this task.

Performance quantification employed both binary and graded scoring
systems tailored to each task's characteristics. Taxonomic
classification used exact match rates for binary assessment alongside
weighted accuracy scoring that awarded partial credit (0.5) for
hierarchically consistent but incomplete answers, such as correct Order
with incorrect Family. The distractor type effect was quantified by
comparing accuracy between phylogenetic and random variants.

Red List category assessment (Task 2) employed exact string matching to
evaluate predictions against ground truth categories. Because only one
categorical label can be correct, accuracy and exact match rate are
equivalent for this task. Additionally, we calculated category
distance---the ordinal steps between predicted and actual
categories---to quantify prediction errors. We also analyzed confusion
patterns between specific category pairs to identify systematic biases
in model predictions.

For geographic distribution (Task 3) and threat identification (Task 4),
the evaluation centered on set-based metrics. A simple exact match is
insufficient for tasks that require identifying a correct set of items
(whether countries or threats), as it fails to credit partially correct
responses. Therefore, this study employed metrics capable of providing a
more nuanced assessment of model performance.

For geographic distribution (Task 3), I used several metrics to quantify
prediction quality. The exact match rate required the predicted set of
countries to be identical to the reference data. To measure partial
correctness, standard information retrieval metrics---precision, recall,
F1 score, and Jaccard similarity were calculated. Additionally, the
invalid response rate was measured the invalid response rate to capture
instances where models generated country names outside the provided
standardized list, indicating a failure to adhere to constraints.

The evaluation of threat identification (Task 4) followed a parallel
approach. While the pool of potential threat categories is limited (n =
12), species often face a specific combination of multiple threats,
making an exact match for the entire profile a highly stringent
criterion. Consequently, in addition to precision, recall, and F1 score,
the mean number of false positives per species was computed. This latter
metric was designed to quantify the systematic over-attribution of
generic threat categories that was observed across all models. All
metrics were computed at species level before macro-averaging, with
standard deviations reported to characterize inter-species variability.
Given the large sample size (n = 21,955) confidence intervals would be
misleadingly narrow, hence the preference for reporting variability
through standard deviations.

\subsection{Evaluation Framework}\label{evaluation-framework}

All evaluation tasks were implemented and executed using Inspect AI
(\citeproc{ref-UK_AI_Security_Institute_Inspect_AI_Framework_2024}{AI
Security Institute 2024}), an open-source framework designed for
reproducible large language model assessments. Inspect AI provided
standardized prompting, logging, and output handling across both
API-based and locally deployed models. This framework allowed for
consistent control of inference settings and systematic capture of model
responses for subsequent scoring.

All models were configured with standardized parameters (temperature =
0.0, top-p = 1.0) to achieve deterministic outputs. For Grok 3, the web
search feature was explicitly disabled to ensure all models were
evaluated under equivalent conditions---relying solely on their
pre-trained knowledge without external information retrieval during
inference. Maximum token limits followed Inspect AI framework defaults,
varying by provider: GPT-4.1 (32,768 tokens), Claude Sonnet 4 (64,000
tokens), Grok 3 (131,072 tokens), while local models via Ollama (Gemma
3-27b and Llama 3.3) used framework defaults (4,096 tokens). All
evaluations were conducted through the Inspect AI framework (v0.3.116)
during August 2025, with version numbers and access timestamps recorded
for reproducibility.

Evaluation metrics were computed using both built-in and custom scorers.
All evaluation logs were stored in JSON format to promote transparency
and reproducibility, and automatic retry mechanisms were employed to
increase robustness against transient API failures. The framework's
caching system further prevented redundant API calls for identical
queries, which significantly reduced evaluation time and associated
costs.

Prompt design followed a structured approach that combined role
specification, output formatting, and sequential instructions. For
instance, in Task 1 the model was explicitly defined as a biological
taxonomy expert, required to produce output in JSON format with specific
fields, and instructed to provide the classification before selecting
the final answer. This approach represented a deliberate balance between
ecological validity and practical necessity.

\subsection{Statistical Analysis}\label{statistical-analysis}

All statistical analyses were conducted in R (version 4.5.1)
(\citeproc{ref-r2025}{R Core Team 2025}). The primary analytical
framework involved generalized linear mixed models (GLMMs) to assess
overall performance, supplemented by specific methods for taxonomic
group comparisons and robustness checks as detailed in the following
subsections.

\subsubsection{Generalized Linear Mixed Models
(GLMMs)}\label{generalized-linear-mixed-models-glmms}

Generalized linear mixed models (GLMMs) were employed to evaluate the
performance of large language models across different tasks while
accounting for the hierarchical structure of the data. A binomial error
distribution with a logit link function was used, as the primary
outcomes (e.g., correctness of responses) were coded as binary
variables.

To account for repeated measures arising from species that were
reassessed multiple times during the study period, species identity was
included as a random effect, controlling for the non-independence of
records within the same species.

Fixed effects included the language model and task type, enabling
comparisons across models and evaluation tasks. Where relevant,
interaction terms between model and task were tested to assess whether
certain tasks disproportionately influenced model performance. Model
selection was guided by Akaike's Information Criterion (AIC), and
residual diagnostics were conducted to ensure the appropriateness of
model fit.

This GLMM framework provided a flexible and statistically robust means
of assessing systematic differences among models while accounting for
repeated observations within species.

Two complementary GLMMs were fitted:

\textbf{Primary Model (Task and Model Effects)}:

\[
logit(\pi_{ij}) = \beta_0 + \beta_1 \times Model_i + \beta_2 \times Task_j + u_{species} + \epsilon_{ij}
\]

\textbf{Taxonomic Analysis Model (Species-level)}:

\[
logit(\pi_{ijk}) = \beta_0 + \beta_1 \times Model_i + \beta_2 \times Task_j + \beta_3 \times TaxonomicGroup_k
               + u_{species} + u_{obs} + \epsilon_{ijk}
\]

where \(\pi\) represents the probability of correct response;
\(\beta_0\) is the intercept; \(\beta_1\), \(\beta_2\), and \(\beta_3\)
are fixed effect coefficients; \(u_{species}\) is a species-level random
effect accounting for repeated measures; \(u_{obs}\) represents
observation-level random effects; and \(\epsilon\) represents residual
variation. The primary model focuses on overall task and model effects,
while the taxonomic model specifically examines group-level differences.

\subsubsection{Taxonomic Group Comparisons and
Robustness}\label{taxonomic-group-comparisons-and-robustness}

To analyze performance differences among major taxonomic groups,
pairwise comparisons were conducted using the GLMM framework. To address
unequal sample sizes across groups, sampling weights were applied, and
sensitivity analyses were performed using balanced subsampling. The
robustness of group-level mean accuracies was evaluated using
nonparametric bootstrap methods, in which species-level accuracies were
resampled with replacement within groups (\ensuremath{10^{4}}
replicates) and 95\% confidence intervals were derived using the
percentile method. For taxonomic groups represented by a single
aggregate observation, binomial simulation-based bootstrapping was
employed to generate confidence intervals.

\subsubsection{Multiple Comparisons and Effect Size
Analysis}\label{multiple-comparisons-and-effect-size-analysis}

Statistical comparisons focused on the four major taxonomic groups:
vertebrates, invertebrates, plants, and fungi. To control for the
family-wise error rate across multiple model comparisons, the Bonferroni
correction was applied (two-sided \(\alpha\) = 0.05).

Effect sizes were calculated to assess practical significance beyond
statistical significance. For binary outcomes, odds ratios (95\% CI)
were reported. For continuous scores, Cohen's d was computed, with
effect sizes interpreted using Cohen's (1988) benchmarks: small (\(|d|\)
= 0.2), medium (\(|d|\) = 0.5), and large (\(|d|\) = 0.8).

\subsection{Taxonomic Group Analysis}\label{taxonomic-group-analysis}

Given the substantial variation in IUCN assessment priorities and data
availability across taxonomic groups, I conducted a detailed analysis of
performance differences among the major groups (vertebrates,
invertebrates, plants, and fungi). This analysis aimed to determine
whether LLM performance varies systematically across these groups.

\subsubsection{Vertebrate Subgroup
Analysis}\label{vertebrate-subgroup-analysis}

Furthermore, given the dominance of vertebrates in conservation
assessments (\citeproc{ref-cazalis2022}{Cazalis et al. 2022};
\citeproc{ref-caldwell2024}{Caldwell et al. 2024}), I performed a more
detailed subgroup analysis within this phylum. This analysis examined
whether traditional conservation ``flagship'' groups (mammals, birds)
(\citeproc{ref-rosenthal2017}{Rosenthal et al. 2017};
\citeproc{ref-davies2018}{Davies et al. 2018}) show systematically
different LLM performance compared to less well-studied classes
(amphibians, fishes, reptiles).

\subsection{Data and Code
Availability}\label{data-and-code-availability}

The code used for this study, including all scoring algorithms, is
publicly available at the project's GitHub repository
(\url{https://github.com/uribo/iucn-redlist-evals/}). The complete
prompt templates for all evaluation tasks are provided in Appendix A.
The evaluation logs generated during this study are not publicly
available as they contain proprietary data derived from the IUCN Red
List.

\section{Results}\label{sec-results}

\subsection{Overall Performance and Task
Hierarchy}\label{overall-performance-and-task-hierarchy}

Analysis of 21,955 species revealed a consistent performance hierarchy
among tasks. Taxonomic classification achieved the highest mean accuracy
(94.9\% \(\pm\) 5\% SD), followed by threat identification (46.4\%
\(\pm\) 1\% SD), geographic distribution (45.6\% \(\pm\) 6.5\% SD), and
Red List category assessment (27.2\% \(\pm\) 9.2\% SD). All accuracies
are macro-averaged across species; see Methods for partial-credit
definitions.

Model rankings were generally consistent across tasks, with GPT-4.1
achieving the highest overall accuracy (67.8\%). An exception was
observed in the Threat Identification task, where Claude Sonnet 4
slightly outperformed other models (Table 2).

\begin{table}[h!]

\caption{\label{tbl-2}Summary of Mean Performance Metrics Across All
Models and Tasks.}

\centering{

\fontsize{6.0pt}{7.2pt}\selectfont
\begin{tabular*}{\linewidth}{@{\extracolsep{\fill}}lccccll}
\toprule
Task Type & Samples & Accuracy (\%) & \(\pm \text{SD}\) & Exact Match Rate & Best Model & Task-Specific Metrics \\ 
\midrule\addlinespace[2.5pt]
Taxonomic Classification (Phylogenetic Distractors) & 109,810 & 91.8 & 8.1 & 91.1 & gpt-4.1 (96.9\%) & Phylogenetic Distractors \\ 
Taxonomic Classification (Random Distractors) & 109,810 & 98.0 & 2.0 & 97.1 & gpt-4.1 (99.5\%) & Random Distractors \\ 
Red List Category Assessment & 109,755 & 27.2 & 9.2 & 27.2 & gpt-4.1 (42.8\%) & Category Distance: 1.03 \\ 
Geographic Distribution & 109,810 & 45.6 & 6.5 & 20.4 & gpt-4.1 (53.4\%) & Precision: 23.414 | Recall: 20.453 \\ 
Threat Identification & 72,380 & 46.4 & 1.0 & 0.2 & claude-sonnet-4 (47.5\%) & F1: 0.033 | Hallucination: 1.66 \\ 
\bottomrule
\end{tabular*}
\begin{minipage}{\linewidth}
n = 21,955 species. Accuracy is macro-averaged per species; partial credit follows Methods (taxonomy: hierarchical consistency; geographic: per-country set overlap; threats: partial scoring). For Red List Category assessment, accuracy equals exact match rate as only one category can be correct. Metrics: Category distance = ordinal steps(\(\pm\) 1 = adjacent); Jaccard = intersection/union.\\
\end{minipage}

}

\end{table}%

Inter-model variance differed substantially by task type. Taxonomic
classification showed minimal variance (CV: 6.8\%), indicating
convergent capabilities for knowledge retrieval. In contrast, Red List
category assessment exhibited high variance (CV: 33.8\%), suggesting
divergent approaches to ecological reasoning tasks.

Across all models, three systematic error patterns were observed. First,
geographic over-prediction was evident, with a mean precision of 23.4\%,
indicating that 77\% of predicted countries were incorrect. Second, a
clear taxonomic bias was detected, as model performance for vertebrates
exceeded that for invertebrates by 6.5\%, a difference that was
statistically significant (\(p\) \textless{} 0.001). Third, models
consistently displayed threat over-attribution, producing on average 1.7
false threats per species.

\subsection{Statistical Modeling of Performance
Predictors}\label{statistical-modeling-of-performance-predictors}

The primary model, a GLMM with a binomial distribution and logit link
function, was first fitted to characterize associations with performance
variation while accounting for the bounded nature of accuracy rates. The
model included fixed effects for LLM model identity and task type, with
species as a random effect to account for repeated measures across
tasks. Table 3 reports the estimated main effects relative to the
baseline condition (Claude Sonnet 4 performing the Red List Category
task), and model selection based on AIC strongly supported this
specification over the null model (\(\Delta\text{AIC}\) = 71.4, AIC
weight = 1; Table 4).

\begin{table}[h!]

\caption{\label{tbl-3}Primary GLMM Fixed Effects Estimates.}

\centering{

\fontsize{9.0pt}{10.8pt}\selectfont
\begin{tabular*}{\linewidth}{@{\extracolsep{\fill}}lccccc}
\toprule
Term & \(\beta\pm\text{SE}\) & Odds Ratio & 95\% CI & z value & p-value \\ 
\midrule\addlinespace[2.5pt]
Baseline (Claude/Category) & -1.059 \(\pm\) 0.214 & 0.35 & [0.23, 0.53] & -4.95 & < 0.001 \\ 
Grok 3 & 0.151 \(\pm\) 0.226 & 1.16 & [0.75, 1.81] & 0.67 & 0.504 \\ 
Gemma3 27B & -0.603 \(\pm\) 0.226 & 0.55 & [0.35, 0.85] & -2.67 & 0.008** \\ 
Llama3.3 70B & 0.047 \(\pm\) 0.226 & 1.05 & [0.67, 1.63] & 0.21 & 0.835 \\ 
GPT-4.1 & 0.618 \(\pm\) 0.226 & 1.86 & [1.19, 2.89] & 2.73 & 0.006** \\ 
Task: Geographic & 0.836 \(\pm\) 0.225 & 2.31 & [1.48, 3.59] & 3.71 & < 0.001 \\ 
Task: Taxonomy (Phylogenetic) & 3.708 \(\pm\) 0.226 & 40.77 & [26.20, 63.45] & 16.43 & < 0.001 \\ 
Task: Taxonomy (Random) & 5.192 \(\pm\) 0.227 & 179.86 & [115.27, 280.63] & 22.88 & < 0.001 \\ 
Task: Threats & 0.870 \(\pm\) 0.225 & 2.39 & [1.53, 3.71] & 3.86 & < 0.001 \\ 
\bottomrule
\end{tabular*}
\begin{minipage}{\linewidth}
Reference categories: Claude Sonnet 4 (model), Red List Category (task). Significance codes: *** p < 0.001, ** p < 0.01, * p < 0.05\\
\end{minipage}

}

\end{table}%

\begin{table}[h!]

\caption{\label{tbl-4}Model Selection Statistics Based on AIC
Comparison.}

\centering{

\fontsize{9.0pt}{10.8pt}\selectfont
\begin{tabular*}{\linewidth}{@{\extracolsep{\fill}}lccccc}
\toprule
Model & df & AIC & BIC & \(\Delta\text{AIC}\) & AIC weight \\ 
\midrule\addlinespace[2.5pt]
Primary Model (LLM + Task) & 10 & 419.6 & 431.8 & 0.0 & 1.000 \\ 
Null Model (Intercept only) & 2 & 491.1 & 493.5 & 71.4 & 0.000 \\ 
\bottomrule
\end{tabular*}
\begin{minipage}{\linewidth}
AIC = Akaike Information Criterion; BIC = Bayesian Information Criterion;  \(\Delta\text{AIC}\) = difference from best model; df = degrees of freedom\\
\end{minipage}

}

\end{table}%

The GLMM analysis revealed that both task type and model identity
significantly predicted accuracy (see Appendix B for likelihood ratio
tests). Task type emerged as the dominant predictor, with taxonomic
classification tasks showing substantially higher odds of correct
responses compared to the baseline Red List Category task. The taxonomy
task with random distractors demonstrated the most pronounced positive
effect (\(\beta\) = 5.19, OR = 179.9, \(p\) \textless{} 0.001), followed
by taxonomy with phylogenetic distractors (\(\beta\) = 3.71, OR = 40.8,
\(p\) \textless{} 0.001). Geographic distribution and threat
identification tasks also showed significant positive effects (OR = 2.31
and 2.39, respectively, both \(p\) \textless{} 0.001), while the Red
List Category task remained the most difficult.

Among models, GPT-4.1 demonstrated significantly higher odds of correct
responses compared to the baseline Claude Sonnet 4 (\(\beta\) = 0.62, OR
= 1.86, \(p\) = 0.006), while Gemma3 27B showed significantly lower
performance (\(\beta\) = -0.6, OR = 0.55, \(p\) = 0.008). Grok 3 and
Llama3.3 70B did not differ significantly from Claude Sonnet 4 (\(p\) =
0.504 and \(p\) = 0.835, respectively).

Additional analysis of the random effects indicated moderate
species-level variance (\(\sigma^2\) = 0.127, SD = 0.356), suggesting
that performance consistency varies across species even after accounting
for model and task effects. This species-level variation may reflect
differences in data availability, conservation research intensity, or
taxonomic complexity across the evaluated species.

\subsection{Task-Specific Findings}\label{task-specific-findings}

\subsubsection{Taxonomic Classification (Task
1)}\label{taxonomic-classification-task-1}

Taxonomic classification achieved the highest performance, with 94.1\%
exact match rate and 94.9\% accuracy when partial credit was applied.
The phylogenetically informed distractors reduced accuracy to 91.8\%,
while random distractors yielded 98\%---a 6.2 percentage point
difference demonstrating greater difficulty in distinguishing closely
related taxa.

\subsubsection{Red List Category Assessment (Task
2)}\label{red-list-category-assessment-task-2}

Red List category assessment presented the greatest difficulty,
achieving an accuracy of 27.2\%. Although this performance exceeded the
random baselines of 10\% for a uniform distribution, 22.8\% for a
frequency-weighted distribution, and 32.2\% for the majority class
baseline, it remained the lowest among all evaluated tasks. The mean
category distance of 1.03 ordinal steps revealed that while errors were
common, they typically involved predictions within one or two categories
of the correct answer. Error analysis (Figure 1) indicated systematic
confusion between adjacent categories, most notably from EN to VU with
2,862 cases, and between NT and LC with 1,009 and 2,857 cases
respectively. Threatened categories showed lower accuracy, with CR at
20.8\%, EN at 18.2\%, and VU at 24.4\%, compared with non-threatened
categories, which yielded 9\% for NT and 28.5\% for LC. The DD category
achieved the highest accuracy at 49.1\%.

\begin{figure}[H]

{\centering \pandocbounded{\includegraphics[keepaspectratio]{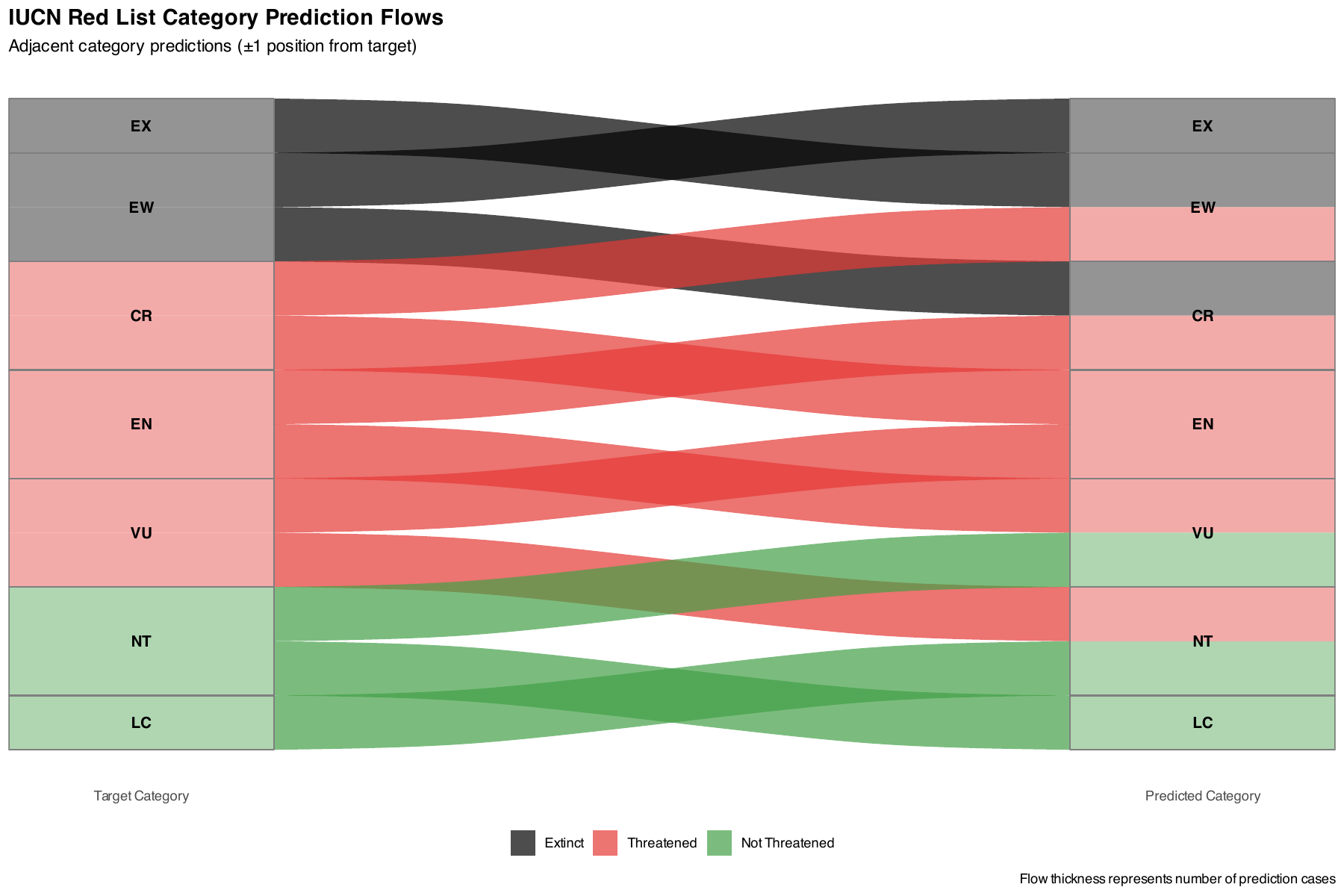}}

}

\caption{Alluvial diagram of IUCN Red List category predictions
aggregated across all evaluated LLMs (n = 109,755 predictions from
21,951 species with valid Red List status). The diagram links target
categories (left) with predicted categories (right), with flow thickness
proportional to the number of predictions. Vertical flows represent
correct classifications, while diagonal flows indicate
misclassifications. Most errors occur between adjacent categories (e.g.,
EN and VU: 2,862 cases; NT and LC: 3,866 cases), indicating that LLMs
tend to confuse neighboring threat levels rather than making random
errors. This suggests that the models capture the ordinal structure of
Red List categories but struggle with applying precise thresholds.}

\end{figure}%

\subsubsection{Geographic Distribution (Task
3)}\label{geographic-distribution-task-3}

Geographic distribution assessment achieved 45.6\% accuracy with
systematic over-prediction: precision 23.4\%, recall 20.5\%, Jaccard
index 20.4\%. Approximately 5\% of responses included invalid country
names.

\subsubsection{Threat Identification (Task
4)}\label{threat-identification-task-4}

Threat identification showed limited set-based performance with
precision 4.2\% and recall 2.9\%. Despite 90.3\% of responses containing
at least one correct threat (contributing to 46.4\% graded accuracy),
exact matches occurred in only 0.2\% of cases. Models averaged 1.7 false
threats per species, demonstrating systematic over-attribution.

\subsection{Evidence of Taxonomic
Bias}\label{evidence-of-taxonomic-bias}

To examine systematic differences across taxonomic groups, a taxonomic
analysis model that included taxonomic group as a fixed effect and both
species-level and observation-level random effects was employed. Figure
2 visualizes group-by-task accuracies as a heatmap.

Estimated variance components were \(u_{obs}\) (\(\sigma^2\) = 0, SD =
0.002) and species intercept (\(\sigma^2\) = 1.101, SD = 1.049), with an
overdispersion statistic (deviance/df = 0.5), which demonstrated
negligible residual overdispersion.

\begin{figure}[H]

{\centering \pandocbounded{\includegraphics[keepaspectratio]{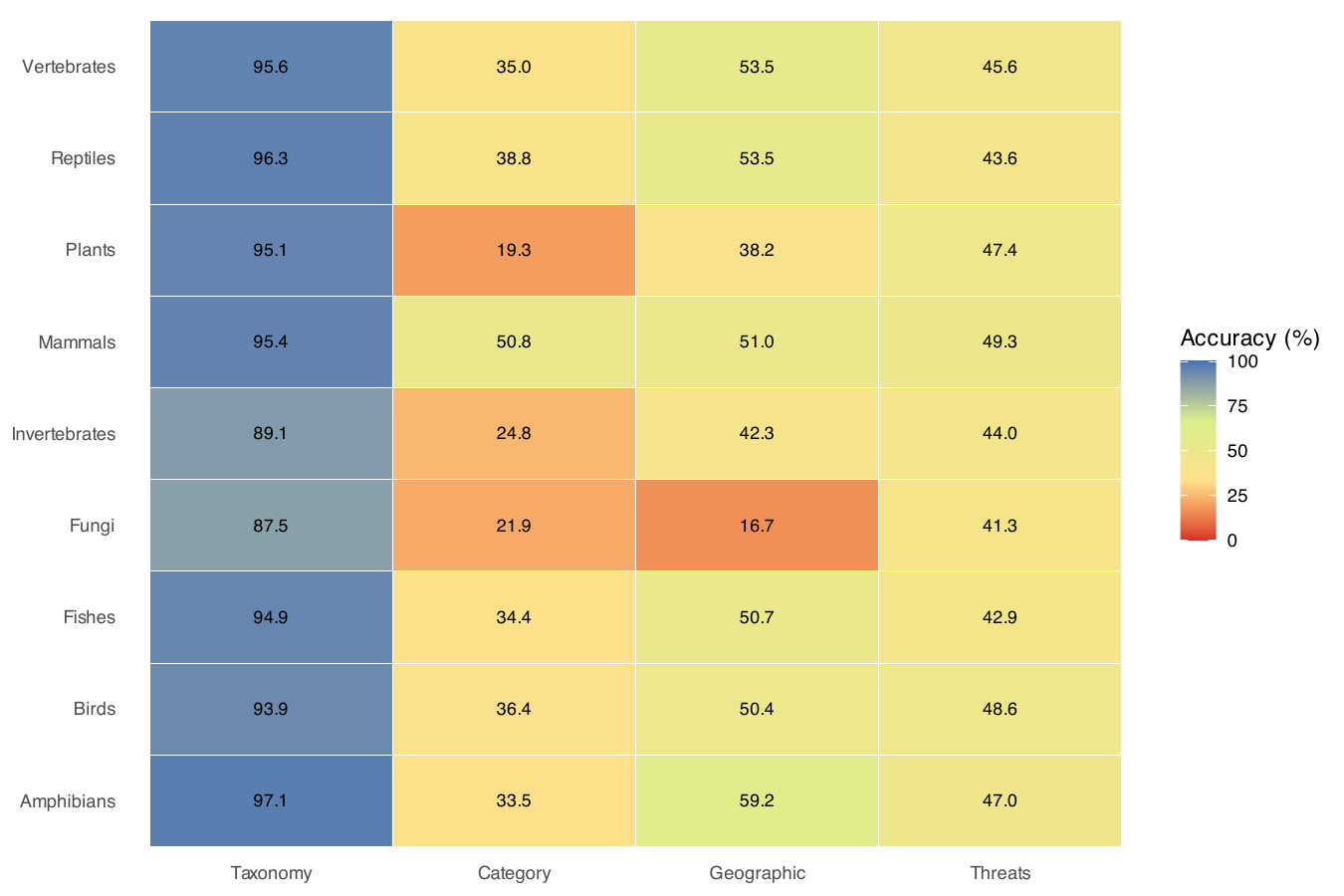}}

}

\caption{Heatmap of task-specific prediction accuracy by taxonomic group
across five LLMs. Cell values indicate mean accuracy (\%) with color
intensity proportional to accuracy. Vertebrates consistently achieve the
highest accuracy across all tasks (mean: 62.3\%), while fungi show the
lowest (mean: 45.8\%). The variation across groups is minimal for
taxonomic classification (range: 87.5--95.6\%) but substantial for
conservation-specific tasks. For instance, in Red List category
assessment, mammals (50.8\%) outperform amphibians (33.5\%) by 17.3
percentage points.}

\end{figure}%

\subsubsection{Taxonomic Performance
Patterns}\label{taxonomic-performance-patterns}

Systematic performance variations were observed across major taxonomic
groups, with vertebrates consistently outperforming other groups (Figure
2). In the taxonomic classification task (Task 1), vertebrates reached
95.6\% accuracy, compared to 89.1\% for invertebrates, 95.1\% for
plants, and 87.5\% for fungi. In the Red List category assessment,
vertebrates achieved 35\% accuracy while plants achieved 19.3\%. In the
geographic distribution task, vertebrates reached 53.5\% compared to
16.7\% for fungi. These results show that differences between groups
were relatively small in basic taxonomic classification but became much
more pronounced in tasks requiring geographic or conservation status
knowledge.

\subsubsection{Vertebrate Class
Analysis}\label{vertebrate-class-analysis}

Within vertebrates, a consistent performance gradient was observed
favoring traditionally well-studied taxa. Amphibians achieved the
highest mean accuracy at 97.1\%, followed by reptiles (96.3\%), mammals
(95.4\%), fishes (94.9\%) and birds (93.9\%). This 3-percentage-point
range within vertebrates is smaller than the between-kingdom differences
but still represents meaningful variation.

The performance gap was most pronounced in Red List category assessment,
where mammals achieved 50.8\% accuracy---substantially higher than birds
(36.4\%), fishes (34.4\%), or amphibians (33.5\%). This represents a
17.3-percentage-point gap between the highest-performing (mammals) and
lowest-performing (amphibians) groups in this task.

Conversely, for geographic distribution, amphibians unexpectedly
outperformed other vertebrate classes with 59.2\% accuracy, compared to
51\% for mammals. This reversal may reflect the more restricted and
well-defined ranges of many amphibian species, making their
distributions easier for models to learn and reproduce accurately.

\subsubsection{Statistical Significance of Taxonomic
Differences}\label{statistical-significance-of-taxonomic-differences}

The taxonomic analysis model confirmed that taxonomic group was a
significant predictor of accuracy (likelihood-ratio test: \(\chi^2\) =
2597.38, \(p\) \textless{} 0.001), explaining 2.1\% of variance after
controlling for task and model effects. The inclusion of taxonomic group
as a fixed effect substantially improved model fit (\(\Delta\) AIC =
8,810 compared to the model without taxonomic group).
Bonferroni-corrected pairwise comparisons were conducted among the four
major groups-vertebrates, invertebrates, plants, and fungi- while
chromista was excluded from the statistical analysis. The adjusted
significance threshold was \(\alpha = 0.05/6 = 0.0083\). Results showed
that vertebrates outperformed the other major groups (all \(p<0.001\)),
with standardized mean differences of \(d\) = 0.2 versus invertebrates
(small effect), \(d\) = 0.02 versus plants (negligible), and \(d\) =
0.27 versus fungi (small effect). These effects persisted in sensitivity
analyses controlling for species name length, geographic range size, and
time since original description.

Sensitivity analyses using weighted samples and balanced subsampling
confirmed the robustness of these findings, with Cohen's \(d\) for the
vertebrate--invertebrate contrast ranging from 0.15 to 0.25 and
taxonomic group variance explained ranging from 1.8\% to 2.4\% across
different sampling strategies.

Table 5 presents the detailed accuracy metrics by taxonomic group for
the Taxonomic Classification task. Bootstrap 95\% confidence intervals
further indicated that vertebrates {[}95.5\%, 95.7\%{]}, showed
significantly higher accuracy than invertebrates {[}88.5\%, 89.6\%{]},
plants {[}94.9\%, 95.2\%{]}, and fungi {[}86.2\%, 88.9\%{]}, with no
overlap between vertebrates and any other group.

\begin{table}[h!]

\caption{\label{tbl-5}Taxonomic Group Performance Statistics}

\centering{

\fontsize{9.0pt}{10.8pt}\selectfont
\begin{tabular*}{\linewidth}{@{\extracolsep{\fill}}lrrrrr}
\toprule
Taxonomic Group & Mean Accuracy & Number of Species & SE & 95\% CI Lower & 95\% CI Upper \\ 
\midrule\addlinespace[2.5pt]
Amphibians & 0.971 & 3,187 & 0.0012 & 0.969 & 0.973 \\ 
Reptiles & 0.963 & 917 & 0.0031 & 0.957 & 0.969 \\ 
Mammals & 0.954 & 193 & 0.0060 & 0.943 & 0.966 \\ 
Plants & 0.951 & 9,800 & 0.0011 & 0.949 & 0.953 \\ 
Fishes & 0.949 & 5,055 & 0.0014 & 0.947 & 0.952 \\ 
Birds & 0.939 & 1,223 & 0.0034 & 0.933 & 0.946 \\ 
Invertebrates & 0.891 & 1,353 & 0.0045 & 0.882 & 0.900 \\ 
\bottomrule
\end{tabular*}
\begin{minipage}{\linewidth}
Mean accuracy for the Taxonomic Classification task (Task 1) with 95\% confidence intervals using group-specific standard errors\\
\end{minipage}

}

\end{table}%

\section{Discussion}\label{sec-discussion}

The results of this study demonstrate a consistent dichotomy in the
performance of large language models: they excel at knowledge-retrieval
tasks but struggle with ecological reasoning. This divergence is
demonstrated by the 67.7-percentage-point gap between taxonomic
classification accuracy (94.9\%) and conservation status assessment
(27.2\%). Because this pattern was consistent across all five models, it
reflects structural task differences rather than deficiencies of
individual architectures.

\subsection{Theoretical Implications: Information Processing
vs.~Judgment
Formation}\label{theoretical-implications-information-processing-vs.-judgment-formation}

This performance dichotomy can be explained by a conceptual framework
distinguishing between information processing and judgment formation.
Information processing involves tasks grounded in stable,
context-independent regularities such as taxonomy, terminology
extraction, or factual summarization. Judgment formation, in contrast,
requires integrating heterogeneous evidence, applying quantitative
thresholds, and reasoning under uncertainty, as in Red List category
assignment or species-specific threat attribution. The 67.7-point gap
between taxonomic classification (zero-shot) and conservation status
assessment (few-shot with examples) reflects a fundamental architectural
limitation that persists despite the provision of examples.
Transformer-based models excel at capturing distributional semantics.
However, they struggle with symbolic reasoning and threshold-based
inference (\citeproc{ref-lin2022}{Lin et al. 2022};
\citeproc{ref-bommasani2022}{Bommasani et al. 2022};
\citeproc{ref-nawaz2025}{Nawaz, Anees-ur-Rahaman, and Saeed 2025}).

Three distinct mechanisms appear to underlie this performance
divergence. First, and most fundamentally, models struggle with
quantitative thresholds that define conservation categories. They
frequently confuse adjacent Red List categories, particularly EN and VU
(2,862 cases, representing 21.2\% of all misclassifications). While
models grasp semantic similarity between categories, they cannot apply
the specific criteria---population decline rates, range restrictions,
and demographic thresholds---that distinguish them. This reflects
transformer models' broader limitation in symbolic reasoning
(\citeproc{ref-sun2024}{Sun et al. 2024}).

Second, the contrast between context-independent and context-dependent
knowledge explains additional variation. Taxonomic classifications
represent stable facts that remain consistent across contexts.
Conservation status and threat identification, however, require
integrating spatiotemporal evidence and regional variation. Models
systematically over-predicted threats (1.7 false threats per species)
and generated invalid country names (77\% of predicted countries were
incorrect) and included invalid country names in about 5\% of responses,
demonstrating how they default to statistically probable outputs rather
than contextually accurate ones.

Third, performance differences across taxa reveal the constraints of
data availability. Models achieved high accuracy for well-documented
vertebrates but showed marked degradation for data-sparse groups. Fungi,
with substantially fewer training examples, showed the lowest accuracy.
This gradient mirrors the uneven distribution of scientific knowledge
across the tree of life, suggesting that model performance is bounded by
training data representation rather than architectural limitations
alone, consistent with long-standing recognition of the profound
underdocumentation of fungi (\citeproc{ref-hawksworth2017}{Hawksworth
and Lücking 2017}).

This dichotomy provides predictive value for conservation tasks.
Information-processing tasks (fixed, well-documented,
context-independent) allow LLMs to approximate expert retrieval, while
judgment-formation tasks (constraint satisfaction, causal inference)
show systematic errors. The observed patterns---adjacent category
confusion (EN/VU: 2,862 cases, 21.2\% of misclassifications), threat
over-attribution (1.7 false threats/species), and taxonomic performance
gradients (Figure 2)---exemplify how models prioritize probabilistic
coherence over ecological correctness.

The framework reveals how model outputs reproduce knowledge asymmetries:
mammals and birds benefit from rich scientific representation, while
fungi and invertebrates suffer from data scarcity. This pattern aligns
with broader observations in biodiversity research: longstanding
syntheses have categorized such gaps as systemic ``shortfalls'' in
biodiversity knowledge (\citeproc{ref-hortal2015}{Hortal et al. 2015}),
and more recent analyses emphasize that taxonomic and geographic biases
remain central challenges to AI-based inference
(\citeproc{ref-pollock2025}{Pollock et al. 2025}).

From a design perspective, the framework supports a division of labor
between AI and human expertise. LLMs excel at scaling literature triage,
extracting candidate threats or range statements, and generating
structured evidence summaries. Castro et al.
(\citeproc{ref-castro2024}{2024}) and Keck, Broadbent, and Altermatt
(\citeproc{ref-keck2025}{2025}) demonstrate successful extraction of
species distributions and ecological interactions from unstructured
texts. However, our results show that challenges intensify when moving
from extraction to evaluative tasks.

Judgment formation must remain human-led, supported by assessor-facing
platforms that formalize Red List calculations
(\citeproc{ref-cazalis2024}{Cazalis et al. 2024}). While techniques like
retrieval-augmented generation may improve specific aspects, they cannot
replace expert judgment for threshold application and causal reasoning.
Recent reviews confirm this limitation: no current method reliably
eliminates hallucinations in high-stakes reasoning
(\citeproc{ref-huang2025}{Huang et al. 2025}), and even hybrid
approaches risk misinformation in complex assessments
(\citeproc{ref-iyer2025}{Iyer et al. 2025}). Human oversight remains
indispensable at critical decision points.

The threat identification task reveals a qualitatively different failure
mode. While the low exact match rate (0.2\%) reflects the inherent
difficulty of predicting precise threat combinations from 12 categories,
the underlying precision (4.2\%) and recall (2.9\%) expose a fundamental
limitation. Models successfully identified at least one correct threat
in 90.3\% of cases, demonstrating surface-level recognition of
conservation concepts. However, they systematically over-attributed
generic threats while missing species-specific factors, producing 1.7
false positives per species alongside severe under-detection. This
pattern suggests that LLMs can retrieve generic threat vocabulary but
cannot perform the ecological reasoning required to map specific species
traits (habitat requirements, life history, distribution) to applicable
threat categories. Unlike Red List categories where adjacent confusion
indicates partial understanding, threat identification requires
multi-dimensional causal inference linking species biology to
anthropogenic pressures---a capability absent in current architectures.
Consequently, while LLMs might assist in generating candidate threat
hypotheses for expert review, they cannot reliably identify threat
profiles without substantive risk of both over-inclusion (diluting
conservation focus) and omission (missing critical threats).

Finally, the theoretical implications extend beyond this case study.
Conservation applications that demand structured extraction of
well-documented information fall within the information-processing
regime, where LLMs can add efficiency with relatively low risk. By
contrast, applications that require synthesizing incomplete evidence or
enforcing explicit criteria fall squarely within the judgment-formation
regime, where uncorrected model outputs risk amplifying knowledge gaps
and automation bias. Explicitly recognizing this boundary conditions the
design of human--AI collaboration: deploying LLMs where they provide
speed and scale, while preserving human oversight where ecological
judgment is indispensable.

\subsection{Reflecting and Amplifying Biases in the Conservation
Knowledge
Base}\label{reflecting-and-amplifying-biases-in-the-conservation-knowledge-base}

These findings are consistent with and contribute to current discourse
on the role of artificial intelligence in conservation. Horizon scans
and commentaries emphasize the potential of AI to accelerate evidence
synthesis and ecological monitoring
(\citeproc{ref-reynolds2024}{Reynolds et al. 2024};
\citeproc{ref-berger-tal2024}{Berger-Tal et al. 2024}), but they also
caution against uncritical techno-optimism, underscoring governance
gaps, uneven benefits, and risks of unintended consequences for
biodiversity (\citeproc{ref-sandbrook2025}{Sandbrook 2025}). Within
biodiversity informatics, LLMs have shown promise for large-scale
information extraction and coding, yet their reliability varies markedly
across tasks and prompts, underscoring the need for domain-grounded
evaluation and continuous human oversight
(\citeproc{ref-farrell2024}{Farrell et al. 2024};
\citeproc{ref-cooper2024}{Cooper et al. 2024};
\citeproc{ref-gougherty2024}{Gougherty and Clipp 2024}). These insights
resonate with the results of the present study: while models performed
strongly on fact retrieval, they were less reliable in tasks requiring
contextual reasoning.

The findings of this study align with and extend the benchmark by Dorm
et al. (\citeproc{ref-dorm2025}{2025}), which showed LLM limitations in
spatial and threat assessment. While confirming their observations
(45.6\% geographic accuracy, 77\% false positive rate), my five-model
evaluation reveals these limitations follow a predictable gradient from
information retrieval to judgment formation---a pattern their two-model
comparison could not detect.

Complementary developments in automated extinction-risk modelling offer
an instructive counterpoint. Predictive pipelines have been used to
infer Red List categories for data-deficient species and to prioritize
reassessments (\citeproc{ref-zizka2021}{Zizka et al. 2021};
\citeproc{ref-caetano2022}{Caetano et al. 2022};
\citeproc{ref-borgelt2022}{Borgelt et al. 2022};
\citeproc{ref-loiseau2024}{Loiseau et al. 2024}). More recently,
assessor-facing platforms such as sRedList have standardized the
derivation of key inputs---including extent of occurrence, area of
occupancy, area of habitat, and population trends---while ensuring
interoperability with the IUCN Species Information System
(\citeproc{ref-cazalis2024}{Cazalis et al. 2024}). These systems
exemplify a pathway to reliability by formalizing evidence calculation,
in contrast to generative models that infer outputs probabilistically
from text.

The present results complement this trajectory by showing that LLMs, if
applied without correction, tend to reproduce existing asymmetries in
the conservation knowledge base. My empirical findings align with
long-standing concerns in the literature regarding such asymmetries.
Mammals and birds, supported by a disproportionately rich textual and
cultural record, are more faithfully represented than groups such as
amphibians, reptiles, invertebrates, and fungi. This taxonomic bias is
not unique to LLMs; broader analyses confirm that cultural perceptions
and media visibility strongly shape the distribution of scientific
effort and conservation investment
(\citeproc{ref-donaldson2017}{Donaldson et al. 2017};
\citeproc{ref-davies2018}{Davies et al. 2018}). These biases extend
beyond funding allocation---taxonomic and geographic imbalances remain
pervasive throughout biodiversity science
(\citeproc{ref-bonnet2002}{Bonnet, Shine, and Lourdais 2002};
\citeproc{ref-amano2013}{Amano and Sutherland 2013};
\citeproc{ref-troudet2017}{Troudet et al. 2017};
\citeproc{ref-caldwell2024}{Caldwell et al. 2024}). Biases also extend
beyond taxonomy: linguistic analyses demonstrate systematic
underutilization of non-English research in global syntheses, where
considerable biodiversity conservation literature published in
non-English languages is overlooked in international assessments
(\citeproc{ref-amano2016}{Amano, González-Varo, and Sutherland 2016};
\citeproc{ref-hannah2025}{Hannah et al. 2025}). Non-English literature
comprises 65\% of references in national biodiversity assessments, yet
international syntheses largely ignore this knowledge
(\citeproc{ref-amano2023}{Amano et al. 2023}). LLMs trained on
English-dominant corpora risk amplifying this bias
(\citeproc{ref-navigli2023}{Navigli, Conia, and Ross 2023};
\citeproc{ref-amano2025}{Amano and Berdejo-Espinola 2025}). Conservation
knowledge from the Global South and non-English regions may become
further marginalized.

Within vertebrates, variation in overall taxonomic classification
accuracy was statistically significant but modest in magnitude, with a
range of only three percentage points between amphibians (97.1\%) and
birds (93.9\%). By contrast, differences became much more pronounced in
Red List category assessments, where mammals achieved 50.8\% accuracy
compared to 33.5\% for amphibians, a gap of over 17.3 percentage points.
This pattern mirrors well-documented biases in conservation research and
funding that disproportionately favor charismatic mammals
(\citeproc{ref-davies2018}{Davies et al. 2018};
\citeproc{ref-cowie2022}{Cowie, Bouchet, and Fontaine 2022}).
Amphibians, while underperforming in conservation status assessments,
achieved the highest accuracy in geographic distribution tasks (59.2\%
versus 51\% for mammals), a result consistent with their often
restricted and ecologically specialized ranges
(\citeproc{ref-wake2008}{Wake and Vredenburg 2008};
\citeproc{ref-luedtke2023}{Luedtke et al. 2023}) that have been
systematically mapped in global assessments. These findings suggest that
while within-vertebrate differences in basic classification are limited,
task-specific biases---particularly in conservation status
assessment---can be substantial and have important implications for
applied use.

In synthesis, these findings indicate that model performance cannot be
understood solely in terms of broad taxonomic divisions but instead
reflects asymmetries in the conservation knowledge base. As visualized
in Figure 2, vertebrates consistently outperform other groups, while
fungi show substantially weaker performance. The 6.5-percentage-point
vertebrate advantage, while statistically modest, could translate to
thousands of misclassified species when applied at scale, and the poor
performance for fungi reveals the fundamental challenge LLMs face when
assessing understudied taxonomic groups. These disparities mirror
long-standing imbalances in biodiversity research and data availability,
where vertebrates, particularly mammals and birds, dominate scientific
output and conservation attention (Titley et al., 2017; Mammola et al.,
2020; Scheffers et al., 2012; (\citeproc{ref-hawksworth2017}{Hawksworth
and Lücking 2017}; \citeproc{ref-troudet2017}{Troudet et al. 2017})).
The observed performance gradient and the species-level variance noted
in my statistical models therefore reflect not intrinsic taxonomic
properties but the uneven distribution and variable quality of
ecological knowledge across the tree of life. Ultimately, my findings
show how LLMs not only reproduce but also risk amplifying existing
inequities in biodiversity science.

\subsection{Limitations and Future
Work}\label{limitations-and-future-work}

Several methodological considerations warrant acknowledgment. First, my
evaluation design prioritized standardization and reproducibility: I
employed zero-shot prompting without optimization, used standardized
taxonomic and geographic vocabularies, and evaluated models with diverse
operational configurations (including Grok 3's real-time search
capability). While this approach ensures comparability and ecological
validity, it may underestimate the performance achievable through
task-specific optimization. Second, my temporal scope---species
reassessed in 2022--2023---provides a robust snapshot but cannot capture
how model performance might evolve with expanded taxonomic coverage or
updated training data. Third, although I identified statistically
significant taxonomic biases, the modest inter-model performance
differences (\textless{} 15 percentage points across most tasks) suggest
that advances will depend more on methodological innovations than
architectural changes.

These findings highlight the need for taxonomically stratified
deployment strategies. Conservation practitioners should apply
group-specific confidence thresholds, with heightened scrutiny for
non-vertebrate assessments. Future development should prioritize
balanced training data across the tree of life rather than architectural
changes alone.

Future work needs to prioritize hybrid system design consistent with the
information-processing versus judgment-formation framework articulated
above. Language models are well suited to scale literature triage,
extract candidate range statements and threats, and produce structured
summaries of evidence. The decisive steps---applying quantitative
thresholds, reconciling conflicting evidence, and assigning Red List
categories---must remain explicitly human, supported by assessor-facing
infrastructures that formalize the calculation of criteria (e.g., extent
of occurrence, area of occupancy, area of habitat, population trends)
and ensure methodological rigor (\citeproc{ref-cazalis2024}{Cazalis et
al. 2024}).

Operationalizing this collaboration requires not only human oversight at
critical decision points but also the development of a consistent
evaluation framework for LLMs in conservation. Such a framework should
be explicitly aligned with the information-processing versus
judgment-formation distinction, ensuring that tasks are assessed with
metrics appropriate to their epistemic demands. Beyond ad hoc
benchmarking, standardized protocols are needed to define task
formulations, specify error taxonomies (e.g., adjacent-category
confusions, generic threat over-attribution, out-of-range localities),
and report taxonomically stratified performance with calibrated
uncertainty. Crucially, this framework must also strengthen connections
with multilingual biodiversity research infrastructures. A substantial
body of conservation evidence exists in non-English sources,
particularly from the Global South, yet these remain systematically
underrepresented in both scientific syntheses and LLM training corpora.
Embedding multilingual data pipelines into evaluation frameworks would
not only reduce linguistic bias but also ensure that LLM-supported
workflows capture regionally critical knowledge, making conservation
assessments more globally equitable and inclusive.

By articulating these limitations, delineating an appropriate division
of labor, and advancing a task-aware, standardized evaluation regime,
the field can transition from static benchmarking toward durable,
practice-ready systems. In this trajectory, LLMs enhance the efficiency
and coverage of information processing, while ecological
judgment---especially where thresholds, causality, and risk are at
stake---remains anchored in expert assessment.

\newpage{}

\section*{References}\label{references}
\addcontentsline{toc}{section}{References}

\phantomsection\label{refs}

\newpage{}

\appendix

\section*{Appendix A. Prompt
Templates}\label{appendix-a.-prompt-templates}
\addcontentsline{toc}{section}{Appendix A. Prompt Templates}

This document provides the complete prompt templates used for each
evaluation task in the IUCN Red List LLM assessment project. Each task
employs specialized prompts designed to optimize for accuracy,
consistency, and token efficiency.

\subsection*{Task 1: Taxonomic
Classification}\label{task-1-taxonomic-classification}
\addcontentsline{toc}{subsection}{Task 1: Taxonomic Classification}

\begin{figure}[h]
\centering
\begin{tcolorbox}
[colback=black!5!white,colframe=gray!75!black,title=Task 1 System Message]
\scriptsize
\begin{verbatim}
You are a biological taxonomy expert. Given a species name and taxonomic choices, select the correct classification.

First output the classification as JSON:
```json
{"Kingdom": "...", "Phylum": "...", "Class": "...", "Order": "...", "Family": "..."}
```

Then select the correct answer: ANSWER: [letter]
\end{verbatim}
\end{tcolorbox}
\caption{System message template for Task 1 (Taxonomic Classification). 
This prompt instructs the model to act as a biological taxonomy expert 
and specifies the required output format combining JSON classification 
with multiple-choice answer selection.}
\label{fig:task1-symsg}
\end{figure}

\begin{figure}[h]
\centering
\begin{tcolorbox}
[colback=black!5!white,colframe=gray!75!black,title=Task 1 Question Template (JSON-Only Format)]
\scriptsize
\begin{verbatim}
{question}
{choices}

Output only:
1. JSON: {"Kingdom":"...","Phylum":"...","Class":"...","Order":"...","Family":"..."}
2. ANSWER: [letter]
\end{verbatim}
\end{tcolorbox}
\caption{Question template for Task 1 showing the minimal format used 
to reduce token consumption while maintaining task clarity. Variables 
{question} and {choices} are dynamically populated for each species.}
\label{fig:task1-question}
\end{figure}

\subsubsection*{Key Features}\label{key-features}
\addcontentsline{toc}{subsubsection}{Key Features}

\begin{itemize}
\tightlist
\item
  \textbf{Token Optimization}: Uses minimal template that focuses on
  essential information only
\item
  \textbf{Choice Optimization}: Automatically detects common taxonomic
  prefixes and shows only differences
\item
  \textbf{Dual Output}: Requires both JSON classification and multiple
  choice answer
\end{itemize}

\subsection*{Task 2: Red List Category
Assessment}\label{task-2-red-list-category-assessment}
\addcontentsline{toc}{subsection}{Task 2: Red List Category Assessment}

\begin{figure}[h]
\centering
\begin{tcolorbox}
[colback=black!5!white,colframe=gray!75!black,title=Task 2 System Message]
\scriptsize
\begin{verbatim}
You are an expert on the IUCN Red List of Threatened Species. Given a species name (scientific or common), determine its
IUCN Red List category.

IUCN Red List Categories:
- EX: Extinct
- EW: Extinct in the Wild
- CR: Critically Endangered
- EN: Endangered
- VU: Vulnerable
- NT: Near Threatened
- LC: Least Concern
- DD: Data Deficient
- NE: Not Evaluated

Output ONLY the category code (e.g., LC, CR, EN, VU, NT, DD, EX, EW, NE).
If the species is not on the Red List or you cannot determine its status, output: null

Examples:
- Input: "*Ursus arctos*" → Output: LC
- Input: "*Salamandrella keyserlingii*" → Output: LC
- Input: "Unknown species" → Output: null
\end{verbatim}
\end{tcolorbox}
\caption{System message template for Task 2 (Red List Category Assessment) 
defining all nine IUCN categories with their codes and providing examples 
of expected input-output format.}
\label{fig:task2-symsg}
\end{figure}

\subsubsection*{Key Features}\label{key-features-1}
\addcontentsline{toc}{subsubsection}{Key Features}

\begin{itemize}
\tightlist
\item
  \textbf{Explicit Categories}: All 9 IUCN categories with full names
  and codes
\item
  \textbf{Output Format}: Strict requirement for category codes only
\item
  \textbf{Null Handling}: Clear instruction for unknown/unlisted species
\item
  \textbf{Examples}: Concrete input-output pairs for guidance
\end{itemize}

\subsection*{Task 3: Geographic
Distribution}\label{task-3-geographic-distribution}
\addcontentsline{toc}{subsection}{Task 3: Geographic Distribution}

\begin{figure}[h]
\centering
\begin{tcolorbox}
[colback=black!5!white,colframe=gray!75!black,title=Task 3 System Message]
\scriptsize
\begin{verbatim}
You are an expert on species distribution. Given a species name (scientific or common), list ALL countries where this 
species is found.

IMPORTANT INSTRUCTIONS:
- ONLY use country names from the following approved list:
Algeria, Egypt, Libya, Morocco, Tunisia, Western Sahara, Angola, Benin, Botswana, Burkina Faso, Burundi, Cameroon, 
Cabo Verde, Central African Republic, Chad, Comoros, Congo, Congo, The Democratic Republic of the, Côte d'Ivoire, 
Djibouti, Equatorial Guinea [includes the islands of Annobón and Bioko], Eritrea, Eswatini, Ethiopia, Gabon, Gambia, 
Ghana, Guinea, Guinea-Bissau, Kenya, Lesotho, Liberia, Madagascar, Malawi, Mali, Mauritania, Mauritius [includes 
Rodrigues], Mayotte, Mozambique, Namibia, Niger, Nigeria, Réunion, Rwanda, Saint Helena, Ascension and Tristan da Cunha, 
Sao Tome and Principe, Senegal, Seychelles [includes the island of Aldabra], Sierra Leone, Somalia, South Africa 
[includes Marion and Prince Edward Islands], South Sudan, Sudan, Tanzania, United Republic of, Togo, Uganda, Zambia, 
Zimbabwe, Bouvet Island, French Southern Territories [includes the Amsterdam-St Paul, Crozet, Kerguelen and Mozambique 
Channel island groups], Heard Island and McDonald Islands, South Georgia and the South Sandwich Islands, China, Hong Kong, 
Japan, Korea, Democratic People's Republic of, Korea, Republic of, Macao, Mongolia, Taiwan, Province of China, Belarus, 
Moldova, Republic of, Russian Federation, Ukraine, Afghanistan, Armenia, Azerbaijan, Bahrain, Cyprus, Georgia, Iran, 
Islamic Republic of, Iraq, Israel, Jordan, Kazakhstan, Kuwait, Kyrgyzstan, Lebanon, Oman, Pakistan, Palestine, 
State of, Qatar, Saudi Arabia, Syrian Arab Republic, Tajikistan, Türkiye, Turkmenistan, United Arab Emirates, Uzbekistan, 
Yemen [includes the island of Socotra], Bangladesh, Bhutan, British Indian Ocean Territory [includes the Chagos 
Archipelago], Brunei Darussalam, Cambodia, Disputed Territory [includes the Paracel Islands and Spratly Islands], India 
[includes the Andaman, Laccadive and Nicobar island groups], Indonesia, Lao People's Democratic Republic, Malaysia, 
Maldives, Myanmar, Nepal, Philippines, Singapore, Sri Lanka, Thailand, Timor-Leste, Viet Nam, Åland Islands, Albania, 
Andorra, Austria, Belgium, Bosnia and Herzegovina, Bulgaria, Croatia, Czechia [was Czech Republic], Denmark, Estonia, 
Faroe Islands, Finland [excludes the Åland Islands], France [includes Clipperton Island in the eastern Pacific Ocean], 
Germany, Gibraltar, Greece, Greenland, Guernsey, Holy See (Vatican City State), Hungary, Iceland, Ireland, Isle of Man, 
Italy, Jersey, Latvia, Liechtenstein, Lithuania, Luxembourg, Malta, Monaco, Montenegro, Netherlands, North Macedonia, 
Norway, Poland, Portugal [includes the Azores, Madeira and the Selvagens islands], Romania, San Marino, Serbia, Slovakia, 
Slovenia, Spain [includes the Belearic and Canary islands and the Spanish North African Territories], Svalbard and 
Jan Mayen, Sweden, Switzerland, United Kingdom of Great Britain and Northern Ireland [excludes Guernsey, Jersey and 
Isle of Man], Anguilla, Antigua and Barbuda, Aruba, Bahamas, Barbados, Bermuda, Cayman Islands, Bonaire, Sint Eustatius 
and Saba, Cuba, Dominica, Curaçao, Dominican Republic, Grenada, Guadeloupe, Haiti, Jamaica, Martinique, Montserrat, 
Puerto Rico, Saint Bathélemy, Saint Kitts and Nevis, Saint Lucia, Saint Martin (French Part), Saint Vincent and the 
Grenadines, Sint Maarten (Dutch Part), Trinidad and Tobago, Turks and Caicos Islands, Virgin Islands, British, Virgin 
Islands, U.S., Belize, Costa Rica, El Salvador, Guatemala, Honduras, Mexico, Nicaragua, Panama, Canada, Saint Pierre and 
Miquelon, United States of America, Argentina, Bolivia, Plurinational State of, Brazil, Chile [includes Easter Island], 
Colombia, Ecuador [includes the Galápagos islands], Falkland Islands (Malvinas), French Guiana, Guyana, Paraguay, Peru, 
Suriname, Uruguay, Venezuela, Bolivarian Republic of, American Samoa, Australia [includes the island groups of 
Ashmore-Cartier, Lord Howe and Macquarie], Christmas Island, Cocos (Keeling) Islands, Cook Islands, Fiji, French Polynesia 
[includes the island groups of the Marquesas, Society, Tuamotu and Tubai], Guam, Kiribati [includes the Gilbert, Kiribati 
Line and Phoenix island groups], Marshall Islands, Micronesia, Federated States of, Nauru, New Caledonia, New Zealand 
[includes the Antipodean, Chatham and Kermadec island groups], Niue, Norfolk Island, Northern Mariana Islands, Palau, 
Papua New Guinea [includes the Bismarck Archipelago and the North Solomons], Pitcairn, Samoa, Solomon Islands, Tokelau, 
Tonga, Tuvalu, United States Minor Outlying Islands [includes the Howland-Baker, Johnston, Midway, US Line and Wake 
island groups], Vanuatu, Wallis and Futuna
- If a species is found in a territory or region not in the approved list, do not include it
- Do NOT include any bracket notations `[` `]` or comments in brackets in your output
- Use only the base country name without any additional notes or clarifications

Output the countries as a semicolon-separated list.

EXAMPLE:
- Input: "*Salamandrella keyserlingii*" → Output: China; Japan; Kazakhstan; Mongolia; Russian Federation
- Input: "*Gentiana pneumonanthe*" → Output: Albania; Bulgaria; Croatia; France; Italy; Montenegro; Portugal; Serbia; 
Slovenia; Spain

Only output the country names from the approved list above, nothing else.
\end{verbatim}
\end{tcolorbox}
\caption{System message template for Task 3 (Geographic Distribution) 
containing the complete list of 195 standardized country names from 
ISO-3166-1 and instructions for semicolon-separated output format.}
\label{fig:task3-symsg}
\end{figure}

\subsubsection*{Key Features}\label{key-features-2}
\addcontentsline{toc}{subsubsection}{Key Features}

\begin{itemize}
\tightlist
\item
  \textbf{Approved Country List}: Dynamic loading of \textasciitilde195
  standardized country names
\item
  \textbf{Bracket Handling}: Explicit instructions to exclude
  territorial annotations
\item
  \textbf{Format Specification}: Semicolon-separated output format
\item
  \textbf{Validation}: Only countries from approved list are accepted
\end{itemize}

\subsubsection*{Country Normalization}\label{country-normalization}
\addcontentsline{toc}{subsubsection}{Country Normalization}

The system includes sophisticated country name normalization:

\begin{itemize}
\tightlist
\item
  Standardizes formal and common names (e.g., ``Bolivia, Plurinational
  State of'' is normalized to ``Bolivia'')
\item
  Removes bracketed expressions to improve matching consistency
\item
  Recognizes common abbreviations and alternative designations
\end{itemize}

\subsection*{Task 4: Threat Assessment}\label{task-4-threat-assessment}
\addcontentsline{toc}{subsection}{Task 4: Threat Assessment}

\begin{figure}[h]
\centering
\begin{tcolorbox}
[colback=black!5!white,colframe=gray!75!black,title=Task 4 System Message]
\scriptsize
\begin{verbatim}
You are an expert on wildlife conservation and extinction risk assessment. Given a species name, identify the primary 
threats that contribute to its extinction risk.

Standard IUCN threat categories:
- Residential & commercial development
- Agriculture & aquaculture
- Energy production & mining
- Transportation & service corridors
- Biological resource use
- Human intrusions & disturbance
- Natural system modifications
- Invasive & other problematic species, genes & diseases
- Pollution
- Geological events
- Climate change & severe weather
- Other options

Output the threat categories as a semicolon-separated list.

Examples:
- Input: "*Salamandrella keyserlingii*" → Output: Residential & commercial development; Energy production & mining, 
Transportation & service corridors, Natural system modifications; Pollution
- Input: "*Gentiana pneumonanthe*" → Output: Agriculture & aquaculture; Natural system modifications; Pollution

Only output the threat category names, nothing else.
\end{verbatim}
\end{tcolorbox}
\caption{System message template for Task 4 (Threat Assessment) listing 
the 12 Level-1 IUCN Threat Categories (Version 3.3) with examples of 
expected output format.}
\label{fig:task4-symsg}
\end{figure}

\subsubsection*{Key Features}\label{key-features-3}
\addcontentsline{toc}{subsubsection}{Key Features}

\begin{itemize}
\tightlist
\item
  \textbf{IUCN Categories}: Based on official IUCN Threats
  Classification Scheme v3.3
\item
  \textbf{Normalization}: Handles `\&' vs `and' variations consistently
\item
  \textbf{Flexible Scoring}: Two scoring approaches available (strict
  vs.~semantic matching)
\end{itemize}

\section*{Appendix B. GLMM Statistical Analysis
Details}\label{appendix-b.-glmm-statistical-analysis-details}
\addcontentsline{toc}{section}{Appendix B. GLMM Statistical Analysis
Details}

This appendix provides full details of GLMM specification, diagnostics,
and post-hoc analyses, complementing the summary results reported in the
main text (Section 3.2)

\subsection*{Model Selection and Likelihood Ratio
Tests}\label{model-selection-and-likelihood-ratio-tests}
\addcontentsline{toc}{subsection}{Model Selection and Likelihood Ratio
Tests}

\subsubsection*{Model Comparison}\label{model-comparison}
\addcontentsline{toc}{subsubsection}{Model Comparison}

\begin{table}[h!]

\caption{\label{tbl-6}Model Selection Statistics for Nested Model
Comparison}

\centering{

\fontsize{9.0pt}{10.8pt}\selectfont
\begin{tabular*}{\linewidth}{@{\extracolsep{\fill}}lcccccc}
\toprule
Model & df & AIC & BIC & logLik & \(\Delta\text{AIC}\) & weight \\ 
\midrule\addlinespace[2.5pt]
M1: Primary (Task + Model) & 10 & 419.6 & 431.8 & -199.8 & 0.0 & 1.0000 \\ 
M0: Null (Intercept only) & 2 & 491.1 & 493.5 & -243.5 & 71.4 & 0.0000 \\ 
\bottomrule
\end{tabular*}
\begin{minipage}{\linewidth}
\(\Delta\text{AIC}\) = difference from best model; weight = Akaike weight.
The Primary Model was selected based on AIC and parsimony.\\
\end{minipage}

}

\end{table}%

\subsubsection*{Likelihood Ratio Tests}\label{likelihood-ratio-tests}
\addcontentsline{toc}{subsubsection}{Likelihood Ratio Tests}

\begin{table}[h!]

\caption{\label{tbl-7}Likelihood Ratio Tests Using Type II Wald
Statistics}

\centering{

\fontsize{9.0pt}{10.8pt}\selectfont
\begin{tabular*}{\linewidth}{@{\extracolsep{\fill}}lccc}
\toprule
Test & \(\chi^2\) & df & p-value \\ 
\midrule\addlinespace[2.5pt]
Null vs. Primary Model & 87.43 & 8 & < 0.001 \\ 
Task Effect & 763.43 & 4 & < 0.001 \\ 
Model Effect & 29.80 & 4 & < 0.001 \\ 
\bottomrule
\end{tabular*}
\begin{minipage}{\linewidth}
Significance codes: *** p < 0.001, ** p < 0.01, * p < 0.05.
    Both task type and model identity significantly predict accuracy.\\
\end{minipage}

}

\end{table}%

\subsection*{Model Diagnostics}\label{model-diagnostics}
\addcontentsline{toc}{subsection}{Model Diagnostics}

\subsubsection*{Convergence and
Optimization}\label{convergence-and-optimization}
\addcontentsline{toc}{subsubsection}{Convergence and Optimization}

The Primary Model converged successfully using the \textbf{bobyqa}
optimizer. Convergence was confirmed by:

\begin{itemize}
\tightlist
\item
  Convergence code: 0 (successful)
\item
  Optimizer: bobyqa (bound optimization by quadratic approximation)
\item
  Function evaluations: 287
\end{itemize}

\subsubsection*{Variance Components and
Overdispersion}\label{variance-components-and-overdispersion}
\addcontentsline{toc}{subsubsection}{Variance Components and
Overdispersion}

\begin{table}[h!]

\caption{\label{tbl-8}Variance Components for Species and
Observation-level Random Effects.}

\centering{

\fontsize{9.0pt}{10.8pt}\selectfont
\begin{tabular*}{\linewidth}{@{\extracolsep{\fill}}llcc}
\toprule
Model & Random Effect & Variance & SD \\ 
\midrule\addlinespace[2.5pt]
Primary GLMM & Species Intercept & 0.1267 & 0.3560 \\ 
Taxonomic Model & obs\_id & 0.0000 & 0.0023 \\ 
Taxonomic Model & species\_id & 1.1010 & 1.0493 \\ 
\bottomrule
\end{tabular*}
\begin{minipage}{\linewidth}
Taxonomic Model includes observation-level random effect to account for overdispersion.\\
\end{minipage}

}

\end{table}%

\textbf{Intraclass Correlation Coefficient (ICC)}: For the Primary
Model, ICC = 0.037, indicating that approximately 3.7\% of variance in
accuracy is attributable to species-level differences.

\textbf{Overdispersion}: Dispersion ratio (deviance/df) = 0.01,
indicating no material overdispersion. The Taxonomic Model included an
observation-level random effect as a precaution.

\subsubsection*{Multicollinearity
Assessment}\label{multicollinearity-assessment}
\addcontentsline{toc}{subsubsection}{Multicollinearity Assessment}

\begin{table}[h!]

\caption{\label{tbl-9}Generalized Variance Inflation Factors for
Multicollinearity Assessment}

\centering{

\fontsize{9.0pt}{10.8pt}\selectfont
\begin{tabular*}{\linewidth}{@{\extracolsep{\fill}}lcccl}
\toprule
Predictor & GVIF & df & \(\text{GVIF}^{1/(2\times\text{df})}\) & Interpretation \\ 
\midrule\addlinespace[2.5pt]
model & 1.000 & 4 & 1.000 & No concern \\ 
task\_type & 1.000 & 4 & 1.000 & No concern \\ 
\bottomrule
\end{tabular*}
\begin{minipage}{\linewidth}
\(\text{GVIF}^{1/(2\times\text{df})}\) \textgreater{} 2 indicates potential multicollinearity. All values \textless{} 2.\\
\end{minipage}

}

\end{table}%

All predictors showed acceptable multicollinearity levels
(\(\text{GVIF}^{(1/(2\times \text{df}))} < 2\)), indicating no inflation
of standard errors due to predictor interdependence.

\subsubsection*{Residual Diagnostics}\label{residual-diagnostics}
\addcontentsline{toc}{subsubsection}{Residual Diagnostics}

Model assumptions were verified through standard diagnostic procedures:

\begin{itemize}
\tightlist
\item
  \textbf{Linearity}: Binomial logit link function appropriate for
  binary outcomes
\item
  \textbf{Independence}: Conditional on species-level random effects,
  observations are independent
\item
  \textbf{Normality of random effects}: Q-Q plot of species-level random
  effects showed approximate normality
\item
  \textbf{Homoscedasticity}: Scale-location plot showed no systematic
  patterns in residual variance
\end{itemize}

All diagnostic checks indicated that model assumptions were adequately
met.

\subsection*{Fixed Effects Estimates}\label{fixed-effects-estimates}
\addcontentsline{toc}{subsection}{Fixed Effects Estimates}

Fixed effects estimates are reported in the main manuscript (Table 3).
Key findings:

\begin{itemize}
\tightlist
\item
  \textbf{Task Type}: Dominant predictor of accuracy (\(\chi^2\) =
  763.4, \(p\) \textless{} 0.001)
\item
  \textbf{Model Identity}: Significant predictor (\(\chi^2\) = 29.8,
  \(p\) \textless{} 0.001)
\item
  \textbf{Taxonomic Group}: Significant predictor in Taxonomic Analysis
  Model (\(\chi^2\) = 2597.4, \(p\) \textless{} 0.001)
\end{itemize}

Complete coefficient table with odds ratios and 95\% confidence
intervals is provided in the main text.

\subsection*{Post-hoc Pairwise
Comparisons}\label{post-hoc-pairwise-comparisons}
\addcontentsline{toc}{subsection}{Post-hoc Pairwise Comparisons}

\subsubsection*{Key Model Comparisons}\label{key-model-comparisons}
\addcontentsline{toc}{subsubsection}{Key Model Comparisons}

\begin{table}[h!]

\caption{\label{tbl-10}Bonferroni-corrected Pairwise Comparisons for Key
Contrasts}

\centering{

\fontsize{9.0pt}{10.8pt}\selectfont
\begin{tabular*}{\linewidth}{@{\extracolsep{\fill}}lcccc}
\toprule
Contrast & \(\beta\) & Odds Ratio & 95\% CI & p-value \\ 
\midrule\addlinespace[2.5pt]
(anthropic/claude-sonnet-4-20250514) - (ollama/gemma3:27b) & 0.60 & 1.83 & [1.17, 2.84] & 0.058 \\ 
(anthropic/claude-sonnet-4-20250514) - (openai/gpt-4.1) & -0.62 & 0.54 & [0.35, 0.84] & 0.050 \\ 
(grok/grok-3) - (ollama/gemma3:27b) & 0.75 & 2.13 & [1.37, 3.31] & 0.007** \\ 
(grok/grok-3) - (openai/gpt-4.1) & -0.47 & 0.63 & [0.40, 0.98] & 0.236 \\ 
(ollama/gemma3:27b) - (ollama/llama3.3:70b) & -0.65 & 0.52 & [0.34, 0.81] & 0.032 \\ 
(ollama/gemma3:27b) - (openai/gpt-4.1) & -1.22 & 0.29 & [0.19, 0.46] & < 0.001 \\ 
\bottomrule
\end{tabular*}
\begin{minipage}{\linewidth}
Only contrasts involving GPT-4.1 (best) and Gemma3 (worst) shown.
      Full comparison table available in supplementary data repository.\\
\end{minipage}

}

\end{table}%

\subsubsection*{Key Task Comparisons}\label{key-task-comparisons}
\addcontentsline{toc}{subsubsection}{Key Task Comparisons}

All task pairwise comparisons were significant at \(p\) \textless{}
0.001 (Bonferroni-corrected). Key contrasts:

\begin{itemize}
\tightlist
\item
  Taxonomy (Random) vs.~Red List Category: OR = 179.9, \(p\) \textless{}
  0.001
\item
  Taxonomy (Phylogenetic) vs.~Red List Category: OR = 40.8, \(p\)
  \textless{} 0.001
\item
  Geographic vs.~Red List Category: OR = 2.31, \(p\) \textless{} 0.001
\item
  Threats vs.~Red List Category: OR = 2.39, \(p\) \textless{} 0.001
\end{itemize}

\subsection*{Taxonomic Group Analysis}\label{taxonomic-group-analysis-1}
\addcontentsline{toc}{subsection}{Taxonomic Group Analysis}

\subsubsection*{Main Group Comparisons}\label{main-group-comparisons}
\addcontentsline{toc}{subsubsection}{Main Group Comparisons}

\begin{table}[h!]

\caption{\label{tbl-11}Vertebrate-centered Taxonomic Contrasts with
Cohen's d and Significance Levels}

\centering{

\fontsize{9.0pt}{10.8pt}\selectfont
\begin{tabular*}{\linewidth}{@{\extracolsep{\fill}}lccc}
\toprule
Contrast & Cohen's d & Effect Size & p-value \\ 
\midrule\addlinespace[2.5pt]
amphibians / invertebrates & 4.11 & Large & < 0.001 \\ 
fishes / invertebrates & 2.30 & Large & < 0.001 \\ 
animals / invertebrates & 2.27 & Large & < 0.001 \\ 
birds / invertebrates & 1.90 & Large & < 0.001 \\ 
\bottomrule
\end{tabular*}
\begin{minipage}{\linewidth}
Cohen's d: small (\(|d|\) = 0.2), medium (\(|d|\) = 0.5), and large (\(|d|\) = 0.8).
Chromista excluded from analysis (n \textless{} 50).\\
\end{minipage}

}

\end{table}%

\textbf{Sensitivity Analyses}: Results were robust to:

\begin{itemize}
\tightlist
\item
  Leave-one-model-out analysis (CV \textless{} 0.10 for all groups)
\item
  Bootstrap confidence intervals (10,000 iterations, BCa method)
\item
  Sample size thresholds (results stable for n \(\ge\) 50)
\end{itemize}

\end{document}